\documentclass[conference, a4paper]{IEEEtran}

\usepackage[a4paper,
            left=1.6cm,
            right=1.6cm,
            top=2.0cm,
            bottom=4.4cm,
            columnsep=0.7cm  
]{geometry}

\usepackage{setspace}
\usepackage{setspace}
\IEEEoverridecommandlockouts
\usepackage{amsmath,amssymb,amsfonts}
\usepackage{amsthm}
\usepackage{algorithm}
\usepackage{algorithmic}
\usepackage{array}
\usepackage{booktabs}%
\usepackage{cite}
\usepackage{color}
\usepackage{enumitem}
\usepackage{graphicx}
\usepackage{mathrsfs}
\usepackage{multirow}
\usepackage{stfloats}
\usepackage{subfigure}
\usepackage{textcomp}
\usepackage{tabularx}
\usepackage{url}
\usepackage{verbatim}
\begin{document}
\title{NeiGAD: Augmenting Graph Anomaly Detection via Spectral Neighbor Information}

\author{Qing Qing,
        Huafei Huang,
        Mingliang Hou,
        Renqiang Luo,
        Mohsen Guizani
\thanks{Qing Qing and Renqiang Luo are with the College of Computer Science and Technology, Jilin University, Changchun 130012, China (qingqing25@mails.jlu.edu.cn, lrenqiang@jlu.edu.cn).}
\thanks{Huafei Huang is with the School of Computer Science and Information Technology, Adelaide University, SA 5095, Australia (huafei.huang@adelaide.edu.au).}
\thanks{Mingliang Hou is with the Guangdong Institute of Smart Education, Jinan University, Guangzhou 510632, China (hml1989@jnu.edu.cn)}
\thanks{Mohsen Guizani is with the Machine Learning Department, Mohamed Bin Zayed University of Artificial Intelligence, Abu Dhabi, United Arab Emirates (mguizani@ieee.org).}
\thanks{The first two authors contributed equally to this work.}
\thanks{Corresponding author: Mingliang Hou, Renqiang Luo.}}

\markboth{IEEE Transactions on Cybernetics,~Vol.~14, No.~8, May~2025}%
{Shell \MakeLowercase{\textit{et al.}}: A Sample Article Using IEEEtran.cls for IEEE Journals}


\maketitle

\begin{abstract}
Graph anomaly detection (GAD) aims to identify irregular nodes or structures in attributed graphs. 
Neighbor information, which reflects both structural connectivity and attribute consistency with surrounding nodes, is essential for distinguishing anomalies from normal patterns. 
Although recent graph neural network (GNN)-based methods incorporate such information through message passing, they often fail to explicitly model its effect or interaction with attributes, limiting detection performance. 
This work introduces NeiGAD, a novel plug-and-play module that captures neighbor information through spectral graph analysis. 
Theoretical insights demonstrate that eigenvectors of the adjacency matrix encode local neighbor interactions and progressively amplify anomaly signals. 
Based on this, NeiGAD selects a compact set of eigenvectors to construct efficient and discriminative representations. 
Experiments on eight real-world datasets show that NeiGAD consistently improves detection accuracy and outperforms state-of-the-art GAD methods. 
These results demonstrate the importance of explicit neighbor modeling and the effectiveness of spectral analysis in anomaly detection.
Code is available at: https://github.com/huafeihuang/NeiGAD. 
\end{abstract}


\section{Introduction}

\begin{figure}[t]
    \centering
    \subfigure {
    \begin{minipage}[b]{0.4\textwidth}
      \centering
      \includegraphics[width=1\textwidth]{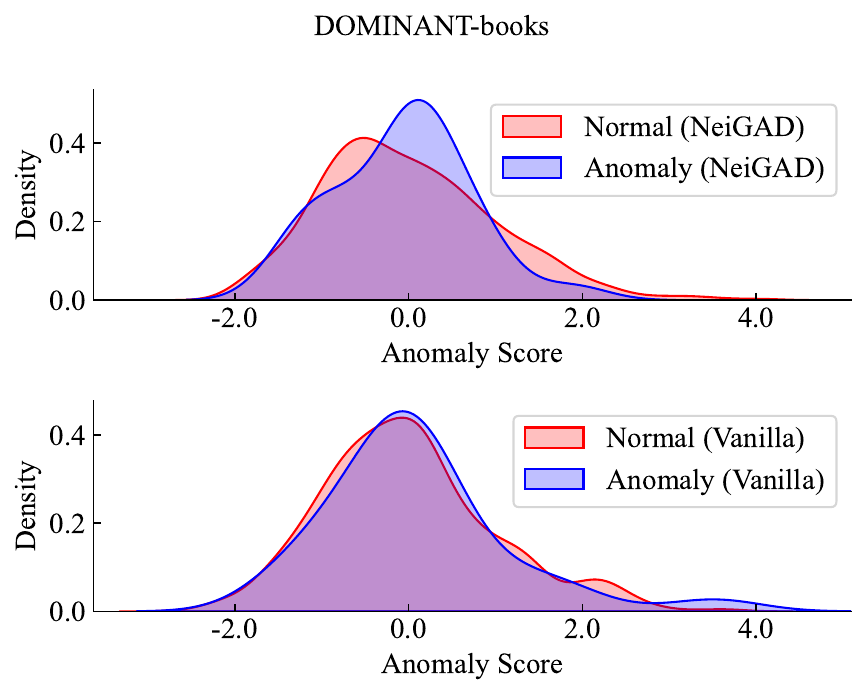}
    \end{minipage}
    }
    \caption{Neighbor information increases the anomaly score gap between normal nodes and anomalous nodes.}
    \label{fig:background}
\end{figure}

\IEEEPARstart{G}raph anomaly detection (GAD) aims to identify atypical graph components (e.g., nodes, edges, or substructures) that significantly deviate from the majority within a graph~\cite{xia2026graph}. 
The increasing interconnectedness of real-world entities and advancements in graph mining techniques have fueled a surge of interest in GAD over the past decade, which has been proven effective in various real-world applications, including financial fraud detection, cyber intrusion detection, device failure prediction, and spam comment identification~\cite{qiao2025deep}.
The field has notably shifted from traditional expert-driven methods to machine learning-based approaches, and more recently, to deep learning techniques, which have significantly improved detection accuracy~\cite{gao2023alleviating}.

\par Anomalous nodes are typically those that substantially deviate from the majority, due to irregular connections (i.e., topology) or inconsistent attributes. 
To detect such deviations, reconstruction error-based autoencoders are widely used in GAD, identifying nodes that are poorly reconstructed based on the local context. 
For example, DOMINANT~\cite{ding2019deep} compresses attributed networks into low-dimensional embeddings and reconstructs both topology and attributes, treating high reconstruction errors as indicators of anomalies. 
Although these methods show some effectiveness, they often treat structural and attribute information separately or combine them in a simplistic way. 
As a result, they fail to fully utilize neighbor information, which captures the consistency between a node and the structure and attributes of its surrounding nodes, thereby limiting performance.

\par Neighbor information, which includes both structural connections and node attributes within a graph, plays a critical role in effective anomaly detection, as anomalies often appear as inconsistencies with the local neighborhood. 
Incorporating such information has been shown to significantly improve GAD performance~\cite{wang2025context}. 
To illustrate this, anomaly scores from a baseline method (DOMINANT~\cite{ding2019deep}) are compared with those from a variant that integrates neighbor information, as shown in Figure~\ref{fig:background}. 
Experimental details and comparisons with other GAD methods are provided in Section~5. 
The observed performance gap highlights neighbor information as a strong additional signal for distinguishing anomalies from normal nodes. 
Although most GNN-based methods propagate neighbor information through message passing, few explicitly model or evaluate its contribution to anomaly detection.

\par Despite its importance, directly modeling neighbor information remains challenging and often incurs high computational cost. 
Recent studies in graph-based tasks suggest that spectral analysis provides an efficient and theoretically grounded alternative. 
For example, FairGT~\cite{luo2024fairgt} demonstrates that spectral truncation can compactly encode neighbor information while reducing computational overhead. 
Similarly, SpecFormer~\cite{bo2023specformer} shows that eigenvectors associated with the largest and smallest eigenvalues effectively encode graph structure. 
This motivates a fundamental question: \textit{How does spectral neighbor information contribute to GAD?} 
To answer this, a theoretical analysis is conducted to uncover the underlying mechanisms.

\par Theoretical analysis yields two key insights regarding spectral information in anomaly detection: 
(1) Adjacency matrix eigenvectors encode the linear average of neighbor components, aligning with message passing to capture local structural information. 
This facilitates the joint modeling of topology and attributes to better distinguish anomalies. 
(2) Utilizing additional eigenvectors linearly amplifies anomaly signals derived from neighbor patterns. 
Consequently, spectral components encode neighborhood interactions and enhance anomaly separability as their dimensionality increases.

\par Based on these insights, we propose a plug-and-play module named NeiGAD (\textbf{Nei}ghbor Information-enhanced \textbf{G}raph \textbf{A}nomaly \textbf{D}etection), designed to improve anomaly detection performance on attributed graphs.
NeiGAD uses the eigenvectors of the adjacency matrix, which capture both connectivity and local structure, to represent neighbor information.
By selecting a subset of these eigenvectors corresponding to specific eigenvalues, NeiGAD captures the most informative patterns while maintaining efficiency.
This approach integrates neighbor-aware signals into existing GAD models with minimal overhead. 
In summary, the contributions are:

\begin{itemize} [leftmargin=0.5cm]
 \item NeiGAD is proposed as a module designed to enhance existing anomaly detection methods by leveraging neighbor information encoded in spectral components.
 \item Eigenvectors inherently encode neighbor information and linearly amplify anomaly signals, offering a principled basis for their use in GAD.
 \item NeiGAD is evaluated on eight real-world datasets, showcasing its effectiveness and superior performance over state-of-the-art GAD methods.
\end{itemize}

\section{Related Work}
\subsection{Graph Anomaly Detection}
\par Graph Auto-Encoders are widely used for GAD by transforming graph data into node embeddings and identifying anomalies via reconstruction quality. 
GAAN~\cite{chen2020generative} employs a generative adversarial framework, detecting anomalies through reconstruction errors and discriminator confidence. 
CoLA~\cite{liu2022anomaly} utilizes contrastive self-supervised learning with instance pair sampling to enhance scalability.
Unlike traditional GNNs that struggle with heterophily, AHFAN~\cite{wang2025graph} introduces a hybrid framework combining spectral filtering and spatial attention to rectify semantic drifts in anomalous nodes.
Finally, GAD-NR~\cite{roy2024gad} focuses on neighborhood reconstruction of local structures and attributes to detect diverse anomalies. 
Despite these advances, most existing methods overlook critical neighbor information, leading to suboptimal performance on real-world datasets.

\subsection{Spectral Graph Learning}
Designing spectral GAD methods necessitates specialized filters, as analyzing anomalies through the lens of the graph spectrum is paramount. 
Specformer~\cite{bo2023specformer} moves beyond traditional scalar-to-scalar filters by introducing a learnable set-to-set spectral filter that employs self-attention to capture global spectrum patterns and non-local dependencies. 
To address inherent biases, FairGT~\cite{luo2024fairgt} integrates a structural feature selection strategy based on adjacency matrix eigenvectors and multi-hop node feature integration, ensuring the independence of sensitive attributes within the Transformer framework. 
Similarly, FairGE~\cite{luo2026fairge} bypasses the risks of sensitive attribute reconstruction by encoding fairness directly through spectral graph theory, utilizing principal eigenvectors to represent structural information. 
Despite these advancements, these spectral filters primarily prioritize global distribution or fairness-aware invariance; they often marginalize intricate local neighbor interrelations and fail to adequately resolve the class and semantic inconsistencies critical for distinguishing anomalies from benign nodes.

\section{Preliminaries}
\subsection{Notations}
\par We denote a set by calligraphic uppercase letters (e.g., $\mathcal{A}$), matrices by bold uppercase letters (e.g., $\mathbf{A}$), and vectors by bold lowercase letters (e.g., $\mathbf{a}$).
A graph is represented as $\mathcal{G} = (\mathcal{V}, \mathbf{A}, \mathbf{X})$, where $\mathcal{V}$ denotes the set of $n$ nodes.
$\mathbf{A} \in \{0, 1\}^{n \times n}$ is the adjacency matrix.
$\mathbf{X} \in \mathbb{R}^{n \times d}$ represents the node feature matrix, where $d$ is the dimension of the node feature.
For a node $v_i$, $\mathcal{N}_i$ denotes its set of neighbor nodes.
For matrix and vector indexing, conventions similar to NumPy in Python are followed.
Specifically, $\mathbf{A}[i,j]$ refers to the element in the $i$-th row and $j$-th column of matrix $\mathbf{A}$, while $\mathbf{A}[i,:]$ and $\mathbf{A}[:,j]$ denote the $i$-th row and the $j$-th column of the matrix, respectively.

\par The adjacency matrix $\mathbf{A}$ spectrum provides a fundamental characterization of the graph's structural topology and connectivity patterns. 
Since $\mathbf{A}$ is a symmetric matrix, its eigendecomposition is given by $\textbf{A} = \mathbf{U}^\top \Lambda \mathbf{U}$, where $\mathbf{U} = (\mathbf{u}_1, \mathbf{u}_2, \dots, \mathbf{u}_n)$ with $\mathbf{u}_i \in \mathbb{R}^{n*1}$.
Here, $\Lambda = diag(\lambda_1, \lambda_2, ......, \lambda_n)$, where $\lambda_i$ is the eigenvalue of $\textbf{A}$, and $\mathbf{u}_i$ is its corresponding eigenvector.
This paper explores how the eigenvector components relate to the anomaly detection.

\subsection{Graph Anomaly Detection}
\par This section introduces the fundamental equations used in GAD, focusing on graph reconstruction methods~\cite{ding2019deep}.
The overall loss function typically combines reconstruction errors from both graph structure and node attributes.
For a GAD based on graph reconstruction, an encoder transforms the node feature matrix $\mathbf{X}$ into a low-dimensional embedding $\mathbf{Z}$ through a series of transformations:
\begin{equation}
    \mathbf{Z} = \textbf{GNN}_{enc}(\mathbf{X},\mathbf{A};\Theta_{enc}).
\end{equation}

\par Here $\textbf{GNN}_{enc}$ represents a GNN-based encoder, and $\Theta_{enc}$ denotes its learnable parameters.
The reconstruction decoders then aim to recover the network structure and attributes from this embedding. 
The structure reconstruction decoder predicts the adjacency matrix $\mathbf{A}'$ using:
\begin{equation}
 \mathbf{A}' = \mathbf{Z}\mathbf{Z}^\top,
\end{equation}
while the attribute reconstruction decoder approximates the attribute matrix $\mathbf{X}'$ as:
\begin{equation}
    \mathbf{X}' = \textbf{GNN}_{dec}(\mathbf{Z},\mathbf{A};\Theta_{dec}).
\end{equation}

\par The loss function, where $\alpha$ is a balancing parameter, is defined as:
\begin{equation}
 \mathcal{L} = (1 - \alpha) \|\mathbf{A} - \mathbf{A}'\|_F^2 + \alpha \|\mathbf{X} - \mathbf{X}'\|_F^2.
\end{equation} 

\par Anomaly scores for nodes are subsequently calculated based on their reconstruction errors, enabling the identification of anomalous nodes as those with higher scores.

\begin{figure*}[t]
	\centering
	\includegraphics[width=0.9\textwidth]{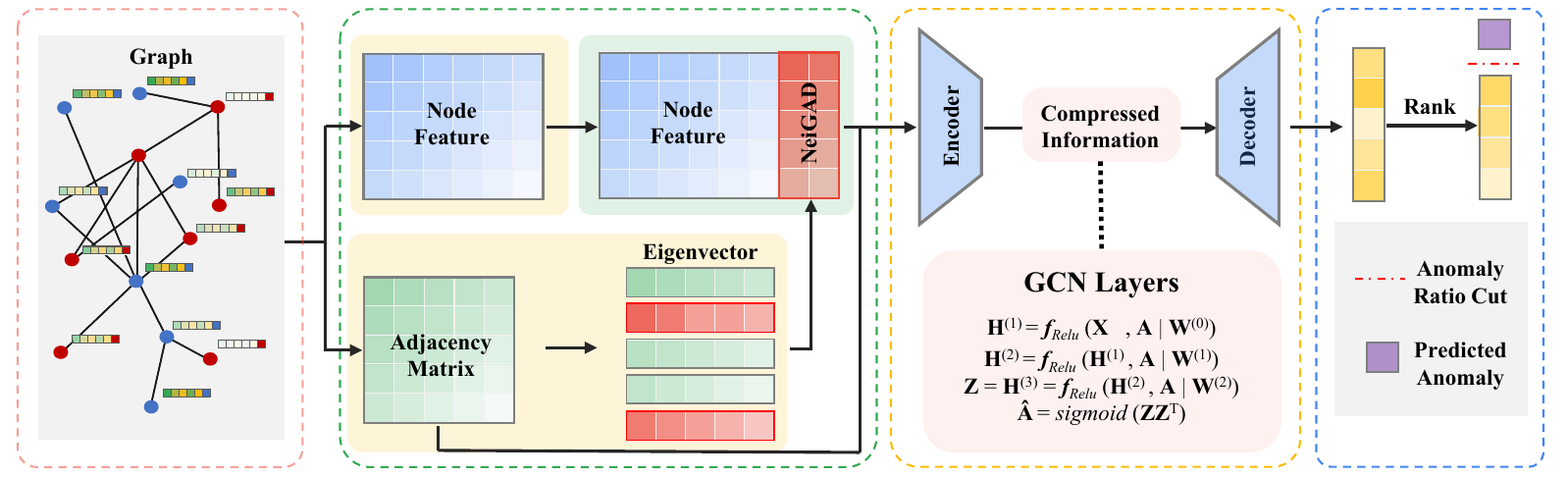}
    \caption{The illustration of NeiGAD.}
    \label{fig:illustration}
\end{figure*}

\section{The Design of NeiGAD}
\par To demonstrate NeiGAD's basic framework, DOMINANT is utilized as a plug-in baseline, as illustrated in Figure~\ref{fig:illustration}. 
Integrating NeiGAD into DOMINANT showcases how this method effectively enhances anomaly detection via spectral neighbor information.

\subsection{Theoretical Findings Underpinning NeiGAD}
\par The neighbor information anomaly score $\mathcal{L}_{\textbf{NI}_i}$ is first defined. 
Normal node attributes resemble those of their neighbors, while abnormal nodes behavior deviates. 
This deviation is termed the neighbor information anomaly degree. 

\textbf{Theorem 1}
\textit{For any eigenvector $\mathbf{u}_i$ of adjacency matrix $\mathbf{A}$, its $j$-th vector component, $\mathbf{u}_{i,j}$, equals the linear average of the components corresponding to node $v_j$ neighbors.}

\textit{Proof.}
\par Assume $\mathbf{u}_i = (\mathbf{u}_{i,1}, \mathbf{u}_{i,2}, \dots, \mathbf{u}_{i,n})^\top$.
Because $\textbf{A} \mathbf{u}_i = \lambda_i \mathbf{u}_i$,thus 
\begin{equation*}
 \begin{aligned}
 \mathbf{u}_{i,j} 
 &=\frac{1}{\lambda_i}\mathbf{A}[j,:]\mathbf{u}_i 
 =\frac{1}{\lambda_i} ( 1 \cdot (\sum_{v_k \in \mathcal{N}_j} \mathbf{u}_{i,k}) \\ &+ 0 \cdot (\sum_{v_k \notin \mathcal{N}_j} \mathbf{u}_{i,k}) )
    =\frac{1}{\lambda_i}(\sum_{v_k \in \mathcal{N}_j} \mathbf{u}_{i,k}).
 \end{aligned}
\end{equation*}

\rightline{$\square$}

\par This theorem reveals an approximate relationship between anomaly detection and the adjacency matrix spectrum.
According to \textbf{Theorem 1}, eigenvector components can be approximated by linearly averaging neighboring nodes's  eigenvector components. 
This property implies that eigenvectors inherently capture spectral neighbor information, referring to the structural and relational characteristics of a node's local neighborhood encoded within the graph's spectral domain.

\textbf{Definition 1}
\textit{Neighbor information anomaly score $\mathcal{L}_{\textbf{NI}_i}$ for node $v_i$ is defined as the reconstruction error of its eigenvector component, where a higher score indicates a higher probability of $v_i$ being anomaly.}
\begin{equation}
 \mathcal{L}_{\textbf{NI}_i} = \|\mathbf{u}_i-\mathbf{u}'_i\|_F^2.
\end{equation}

\par For the eigenvector $\mathbf{u}_i$, the encoder is implicitly defined through iterative multiplication by the adjacency matrix:
\begin{equation}
 \begin{aligned}
 \mathbf{u}_i^{(1)} = \mathbf{A}\mathbf{u}_i, 
 \mathbf{u}_i^{(2)} = \mathbf{A}\mathbf{u}_i^{(1)}, 
 \mathbf{u}_i^{(3)} = \mathbf{A}\mathbf{u}_i^{(2)}.
 \end{aligned}
\end{equation}

\par The decoder is then defined as:
\begin{equation}
    \mathbf{u}'_i = \mathbf{A}\mathbf{u}_i^{(3)}.
\end{equation}

\par This formulation reveals an approximate relationship between $\mathcal{L}_{\textbf{NI}_i}$ and the adjacency matrix's eigenvector components.
Based on this, the objective is to compute the neighbor information anomaly degree using these components.

\textbf{Theorem 2}
\textit{The insertion of eigenvectors linearly increases the neighbor information anomaly degree.
For each positive integer $k$, and for $j \in \{1, 2, \dots k\}$ with $i_j \in \{1, 2, \dots n\}$, where $\mathbf{U}_i = (\mathbf{u}_{i_1}, \mathbf{u}_{i_2} \dots, \mathbf{u}_{i_k})$ denotes selected eigenvectors:}
\begin{equation}
    \mathcal{L}_{\mathbf{X}||\mathbf{U}_{i}} =\mathcal{L}_\mathbf{X} + \mathcal{L}_{\textbf{NI}_{i_1}} + \mathcal{L}_{\textbf{NI}_{i_2}} + \dots + \mathcal{L}_{\textbf{NI}_{i_k}}.
\end{equation}

\textit{Proof.}
\par The theorem is proven by mathematical induction.
For each positive integer $k$, the statement to be proven is:
\begin{equation*}
    \mathcal{L}_{\mathbf{X}||\mathbf{U}_{i}} =\mathcal{L}_\mathbf{X} + \mathcal{L}_{\textbf{NI}_{i_1}} + \mathcal{L}_{\textbf{NI}_{i_2}} + \dots + \mathcal{L}_{\textbf{NI}_{i_k}}.
\end{equation*}

\textbf{Base case} ($\mathbf{k=1}$):
For any $j \in \{1, 2, \dots n\}$, let $i_j \in \{1,2, \dots n\}$.
Consider the combined loss when a single eigenvector $u_{i_1}$ is incorporated with the feature matrix $\mathbf{X}$:
\begin{equation*}
 \begin{aligned}
 & \mathcal{L}_{\mathbf{X}||\mathbf{u}_{i_1}} \\
 =& \Bigl{\Vert} \mathbf{X}||\mathbf{u}_{i_1} - \mathbf{X}'||\mathbf{u}'_{i_1} \Bigl{\Vert}_F^2 \\
 =& \Bigl{\Vert} \mathbf{X} - \mathbf{X}' \Bigl{\Vert}_F^2 + \Bigl{\Vert} \mathbf{u}_{i_1} - \mathbf{u}'_{i_1} \Bigl{\Vert}_F^2 \\
 =& \mathcal{L}_\textbf{X} + \mathcal{L}_{\textbf{NI}_{i_1}}.
 \end{aligned}
\end{equation*}

\par The result holds for $k=1$, validating the base step.

\par \textbf{Inductive Hypothesis} ($\mathbf{k=r}$):
Assume the statement holds true for $k=r$. 
That is, for a set of $r$ eigenvectors $\mathbf{U}_r = (\mathbf{u}_{i_1}, \mathbf{u}_{i_2}, \dots, \mathbf{u}_{i_r})$:
\begin{equation*}
 \mathcal{L}_{\mathbf{X}||\mathbf{U}_r} = \mathcal{L}_\mathbf{X} + \mathcal{L}_{\textbf{NI}_{i_1}} + \mathcal{L}_{\textbf{NI}_{i_2}} + \dots + \mathcal{L}_{\textbf{NI}_{i_r}}.
\end{equation*}

\textbf{Inductive Step} ($\mathbf{k=r+1}$):
Consider the case where an additional eigenvector $u_{i_{r+1}}$ is include, forming $\mathbf{U}_{r+1} = (\mathbf{u}_{i_1}, \mathbf{u}_{i_2}, \dots, \mathbf{u}_{i_r}, \mathbf{u}_{i_{r+1}})$.
The loss function is expressed as:
\begin{equation*}
 \begin{aligned}
 & \mathcal{L}_{\mathbf{X}||\mathbf{U}_{r+1}} \\
 =& \Bigl{\Vert} \mathbf{X}||\mathbf{U}_{r}||\mathbf{u}_{i_{r+1}} - \mathbf{X}'||\mathbf{U}'_{r} || \mathbf{u}'_{i_{r+1}} \Bigl{\Vert}_F^2 \\
 =& \Bigl{\Vert} \mathbf{X}||\mathbf{U}_{r} - \mathbf{X}'||\mathbf{U}'_{r} \Bigl{\Vert}_F^2 + \Bigl{\Vert} \mathbf{u}_{i_{r+1}} - \mathbf{u}'_{i_{r+1}} \Bigl{\Vert}_F^2 \\
 =& \mathcal{L}_\mathbf{X} + \mathcal{L}_{\textbf{NI}_{i_1}} + \mathcal{L}_{\textbf{NI}_{i_2}} + \dots + \mathcal{L}_{\textbf{NI}_{i_r}} + \mathcal{L}_{\textbf{NI}_{i_{r+1}}}.
 \end{aligned}
\end{equation*}

\par It holds for $k=r+1$, confirming the inductive step.

\par By combining the base step, inductive hypothesis, and inductive step, the mathematical statement is proven to hold for all positive integer $k$. 
Therefore, the insertion of eigenvectors has a linearly increasing effect on neighbor information anomaly degree.

\rightline{$\square$}

\par \textbf{Theorem 2} implies that combining multiple eigenvector-derived anomaly scores linearly amplifies anomaly signals.

\subsection{Empirical Observations of NeiGAD}
\par
\begin{figure}[h]
    \centering
    \subfigure {
    \begin{minipage}[b]{0.20\textwidth}
      \centering
      \includegraphics[width=1\textwidth]{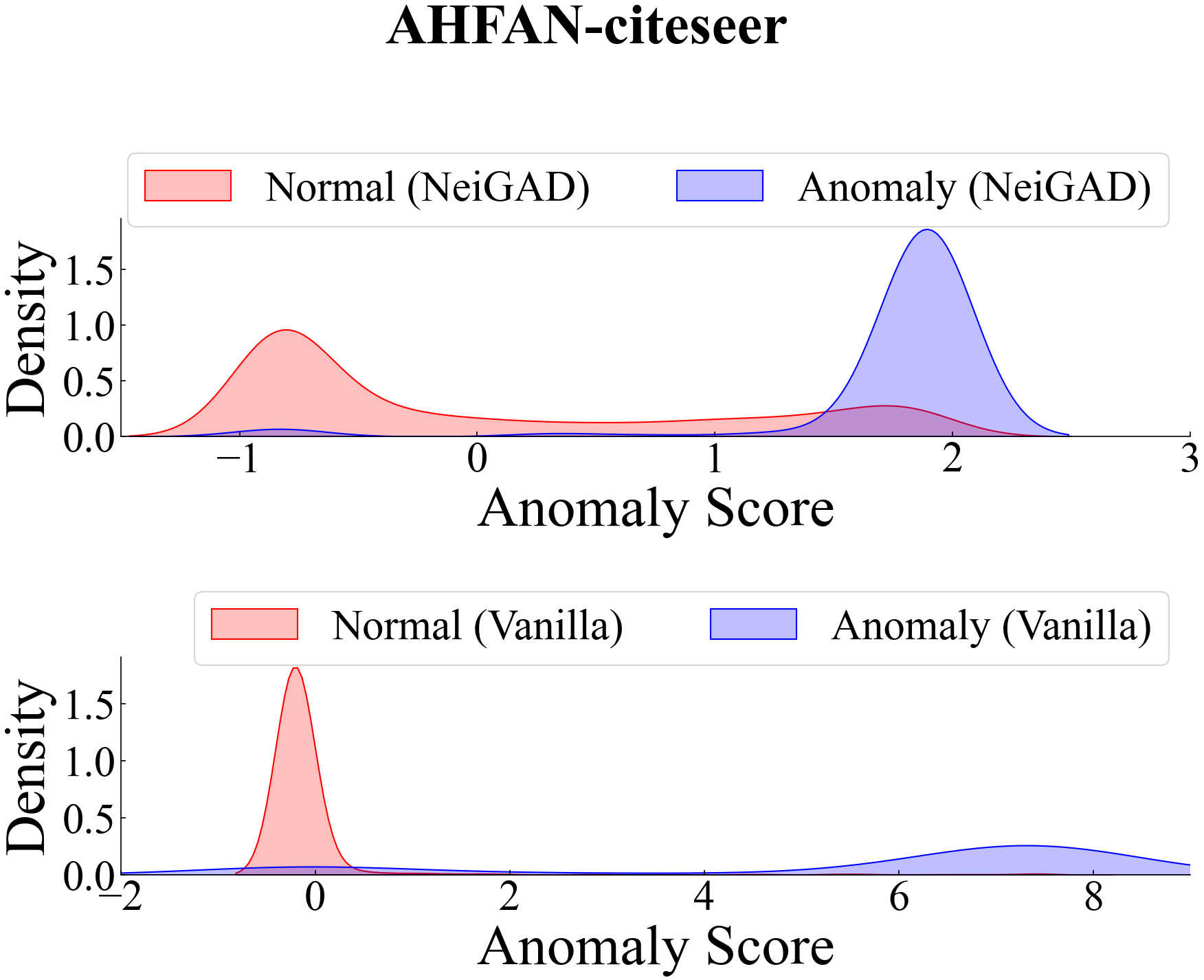}
    \end{minipage}
    }
    \subfigure {
    \begin{minipage}[b]{0.20\textwidth}
      \centering
      \includegraphics[width=1\textwidth]{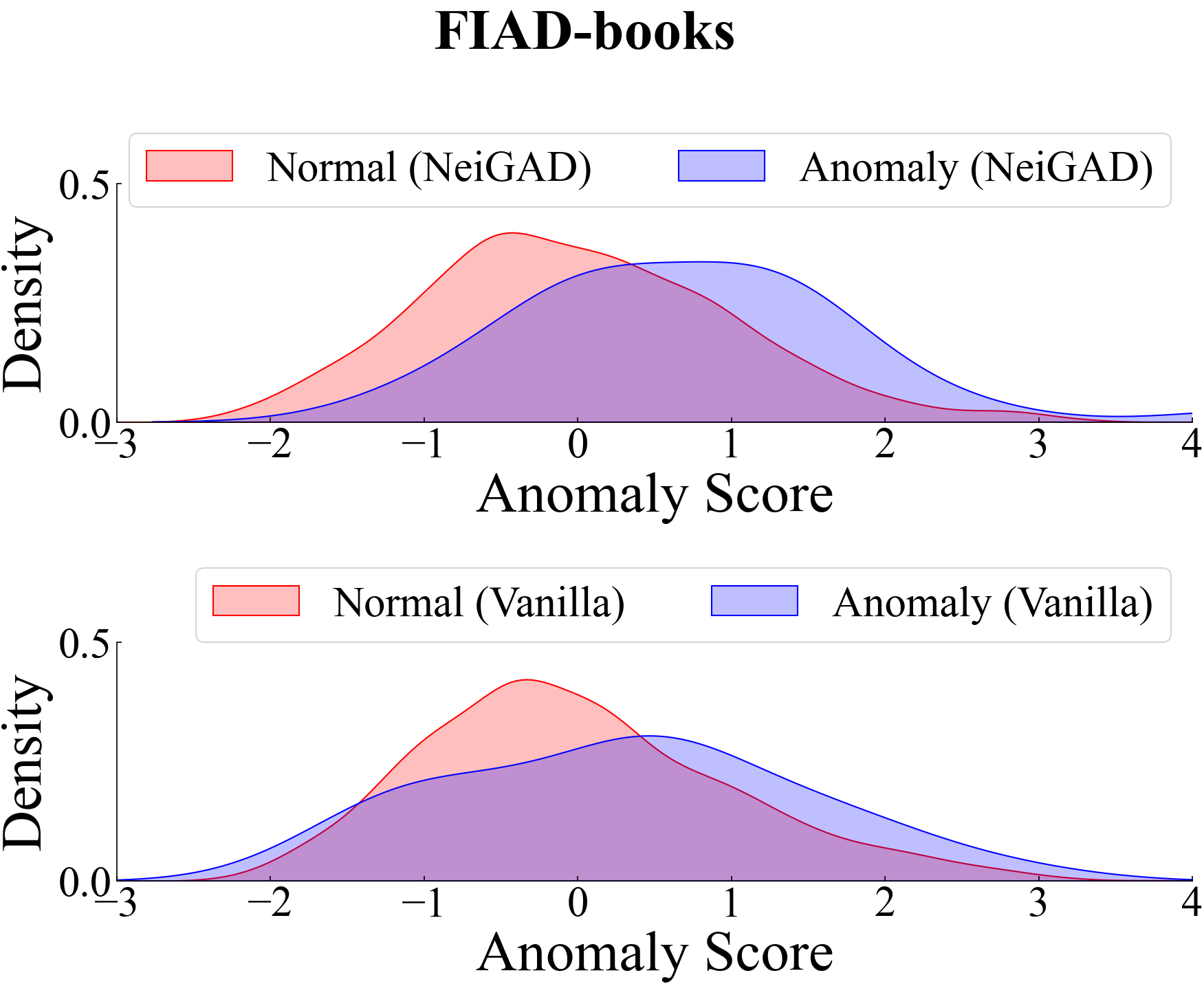}
    \end{minipage}
    }
    \caption{Empirical observation of NeiGAD.}
    \label{fig:EO}
\end{figure}

 The empirical evidence demonstrates that incorporating neighbor information significantly amplifies the discriminative gap between normal and anomalous nodes, making them more distinguishable. 
As illustrated in Figure~\ref{fig:EO}, our analysis examines how these neighbor-based scores, which capture essential node-neighbor relationships, contribute to the total anomaly assessment of individual nodes. 
By evaluating a variety of detection methods across diverse datasets, we provide a comprehensive understanding of how spectral neighbor information improves GAD performance. 
This extensive investigation highlights the general utility of neighbor signals in resolving semantic inconsistencies that traditional methods often overlook.
For further details and extended results, please refer to
https://github.com/huafeihuang/NeiGAD.

\subsection{Technical Details of NeiGAD}
\par NeiGAD extracts spectral information by performing eigendecomposition on the adjacency matrix $\textbf{A}$. 
As established by \textbf{Theorem 1}, the eigenvectors of the adjacency matrix intrinsically encode neighbor information, with each vector component reflecting the linear average of its neighbors. 
To leverage this property, a subset of eigenvectors is selected to represent neighbor information, denoted as:

\begin{equation}
    \mathbf{u}_i = (\mathbf{u}_{i,1}, \mathbf{u}_{i,2}, \dots, \mathbf{u}_{i,n})^\top.
\end{equation}

\par When graph convolution or similar operations are applied, this spectral representation amplifies the mismatch between anomalous nodes and their neighbors, thereby enhancing anomaly detection performance.

\par According to \textbf{Theorem 2}, selecting eigenvector increases the anomaly degree by accentuating the mismatch between nodes and their neighbor information. 
To strike a balance between computational efficiency and detection accuracy, NeiGAD selects $2$-$10$ eigenvectors corresponding to the largest eigenvalues.
This selection is efficiently achieved using the Arnoldi Package algorithm. 
This algorithm directly computes the top $t$ eigenvalues and their corresponding eigenvectors in descending order, circumventing the need for a full eigendecomposition of the adjacency matrix. 
By avoiding the computation of all eigenvectors, NeiGAD significantly reduces computational complexity while maintaining accuracy.

\begin{table*}[htbp]
    \centering
    \caption{Comparison of ROC-AUC in percentage (\%).}
    \footnotesize
    \tabcolsep=0.25cm
    \begin{tabular}{llrrrrrrrr|l}
        \toprule
        \textbf{Method} & \textbf{Version} & \textbf{Books} & \textbf{Enron} & \textbf{Cora} & \textbf{Citeseer} & \textbf{Pubmed} & \textbf{ACM} & \textbf{Blog} & \textbf{DBLP} & Avg \\ 
        \midrule
        \multirow{2}{*}{MLPAE} & vanilla & $68.65$ & $76.37$ & $76.19$ & $73.70$ & $74.60$ & $74.93$ & $77.49$ & $74.39$ & \textcolor{red}{\multirow{2}{*}{$+5.74$}} \\
        & +NeiGAD & \textcolor{red}{$+8.25$} & \textcolor{red}{$+19.44$} & \textcolor{red}{$+0.74$} & \textcolor{red}{$+0.19$} & \textcolor{red}{$+12.54$} & \textcolor{red}{$+3.73$} & \textcolor{red}{$+0.11$} & \textcolor{red}{$+0.89$} \\
        \cmidrule{1-10}
        \multirow{2}{*}{GCNAE} & vanilla & $70.03$ & $74.20$ & $76.36$ & $71.88$ & $74.62$ & $74.81$ & $74.66$ & $74.01$ & \textcolor{red}{\multirow{2}{*}{$+4.08$}} \\
        & +NeiGAD & \textcolor{red}{$+6.10$} & \textcolor{red}{$+6.29$} & \textcolor{red}{$+1.39$} & \textcolor{red}{$+0.96$} & \textcolor{red}{$+12.94$} & \textcolor{red}{$+3.50$} & \textcolor{red}{$+0.28$} & \textcolor{red}{$+1.21$} \\
        \cmidrule{1-10}
        \multirow{2}{*}{DOMINANT} & vanilla & $68.54$ & $71.03$ & $89.79$ & $93.81$ & $83.21$ & $81.44$ & $69.62$ & $96.13$ & \textcolor{red}{\multirow{2}{*}{$+1.26$}} \\
        & +NeiGAD & \textcolor{red}{$+5.49$} & \textcolor{red}{$+3.52$} & \textcolor{red}{$+0.12$} & \textcolor{red}{$+0.45$} & \textcolor{red}{$+0.22$} & \textcolor{red}{$+0.13$} & \textcolor{red}{$+0.08$} & \textcolor{red}{$+0.10$} \\
        \cmidrule{1-10}
        \multirow{2}{*}{AnomalyDAE} & vanilla & $72.52$ & $80.13$ & $77.24$ & $72.12$ & $74.83$ & $74.87$ & $74.80$ & $74.21$ & \textcolor{red}{\multirow{2}{*}{$+3.57$}} \\
        & +NeiGAD & \textcolor{red}{$+3.30$} & \textcolor{red}{$+6.41$} & \textcolor{red}{$+0.67$} & \textcolor{red}{$+0.72$} & \textcolor{red}{$+12.52$} & \textcolor{red}{$+3.53$} & \textcolor{red}{$+0.38$} & \textcolor{red}{$+1.05$} \\
        \cmidrule{1-10}
        \multirow{2}{*}{CoLA} & vanilla & $70.17$ & $77.26$ & $68.87$ & $71.01$ & $56.29$ & $55.96$ & $63.69$ & $55.09$ & \textcolor{red}{\multirow{2}{*}{$+3.26$}} \\
        & +NeiGAD & \textcolor{red}{$+2.05$} & \textcolor{red}{$+12.99$} & \textcolor{red}{$+1.81$} & \textcolor{red}{$+1.64$} & \textcolor{red}{$+2.57$} & \textcolor{red}{$+1.07$} & \textcolor{red}{$+3.18$} & \textcolor{red}{$+0.76$} \\
        \cmidrule{1-10}
        \multirow{2}{*}{GAAN} & vanilla & $67.53$ & $95.11$ & $76.81$ & $73.66$ & $91.89$ & $89.78$ & $80.40$ & $75.17$ & \textcolor{red}{\multirow{2}{*}{$+1.91$}} \\
        & +NeiGAD & \textcolor{red}{$+9.01$} & \textcolor{red}{$+0.49$} & \textcolor{red}{$+0.95$} & \textcolor{red}{$+0.64$} & \textcolor{red}{$+2.37$} & \textcolor{red}{$+1.23$} & \textcolor{red}{$+0.40$} & \textcolor{red}{$+0.19$} \\
        \cmidrule{1-10}
        \multirow{2}{*}{GADNR} & vanilla & $57.93$ & $86.57$ & $77.65$ & $76.72$ & $80.08$ & $73.06$ & $56.52$ & $76.02$ & \textcolor{red}{\multirow{2}{*}{$+2.64$}} \\
        & +NeiGAD & \textcolor{red}{$+5.85$} & \textcolor{red}{$+1.50$} & \textcolor{red}{$+1.50$} & \textcolor{red}{$+3.91$} & \textcolor{red}{$+3.81$} & \textcolor{red}{$+0.99$} & \textcolor{red}{$+3.10$} & \textcolor{red}{$+0.42$} \\
        \cmidrule{1-10}
        \multirow{2}{*}{FIAD} & vanilla & $63.38$ & $61.34$ & $89.22$ & $93.20$ & $89.69$ & $89.52$ & $74.25$ & $96.30$ & \textcolor{red}{\multirow{2}{*}{$+3.30$}} \\
        & +NeiGAD & \textcolor{red}{$+13.75$} & \textcolor{red}{$+5.97$} & \textcolor{red}{$+1.62$} & \textcolor{red}{$+1.45$} & \textcolor{red}{$+2.11$} & \textcolor{red}{$+0.80$} & \textcolor{red}{$+0.42$} & \textcolor{red}{$+0.26$} \\
        \cmidrule{1-10}
        \multirow{2}{*}{AHFAN} & vanilla & $70.94$ & $72.63$ & $87.37$ & $90.25$ & $87.09$ & $87.01$ & $87.47$ & $90.39$ & \textcolor{red}{\multirow{2}{*}{$+4.09$}} \\
        & +NeiGAD & \textcolor{red}{$+2.64$} & \textcolor{red}{$+7.00$} & \textcolor{red}{$+3.18$} & \textcolor{red}{$+2.21$} & \textcolor{red}{$+4.51$} & \textcolor{red}{$+2.89$} & \textcolor{red}{$+6.29$} & \textcolor{red}{$+3.97$} \\
        \cmidrule{1-10}
        & Avg & \textcolor{red}{$+6.27$} & \textcolor{red}{$+7.07$} & \textcolor{red}{$+1.33$} & \textcolor{red}{$+1.35$} & \textcolor{red}{$+5.95$} & \textcolor{red}{$+1.99$} & \textcolor{red}{$+1.62$} & \textcolor{red}{$+0.98$} & \\
        \bottomrule
    \end{tabular}
    \label{tab:comparison}
\end{table*}

\subsection{Complexity of NeiGAD}
\par Eigendecomposition typically incurs $\mathcal{O}(n^3)$ computational complexity for capturing global neighbor information. 
However, NeiGAD optimizes this by avoiding full eigendecomposition or eigenvector ranking.
Instead, it selectively focuses on a small subset of eigenvectors, reducing unnecessary computations. 
Utilizing the ARPACK, NeiGAD achieves a lower time complexity of $\mathcal{O}(n t^2)$, where $t$ is the number of selected eigenvectors, and a space complexity of $\mathcal{O}(n)$. 
This approach ensures computational efficiency and scalability.

\section{Experiments}

\subsection{Datasets}
We selected eight commonly used real-world graph datasets for the experiments.
\begin{itemize} [leftmargin=0.5cm]
    \item \textbf{Books}~\cite{sanchezm2013statistical}: Sourced from the Amazon network, utilizing user-provided tags.
    The outlier ground truth is defined by products tagged as `amazonfail' by $28$ users, reflecting user disagreement with sales ranks.
    \item \textbf{Enron}~\cite{sanchezm2013statistical}: A communication network where email transmissions form edges between email addresses. 
    Each node possesses $18$ attributes, including aggregated statistics on average email content length and recipient count.
    \item \textbf{Cora}~\cite{sen2008collective}, \textbf{Citeseer}~\cite{sen2008collective}, \textbf{Pubmed}~\cite{sen2008collective}, \textbf{ACM}~\cite{tang2008arnetminer}: Citation networks of scientific publications. 
    Nodes represent articles, and edges denote citation relationships. 
    The attribute vector for each node is a bag-of-words representation determined by the dictionary size.
    \item \textbf{BlogCatalog}\textbf{} (shortened to Blog): A social network from the BlogCatalog platform. 
Nodes represent users, and edges indicate follower relationships.
    \item \textbf{DBLP}~\cite{yuan2021higher}: A citation network comprising $5,484$ scientific publications and $8,117$ citation edges. 
\end{itemize}

\subsection{Baselines}
\par The experiments utilize nine commonly used GAD methods as baselines, to which the proposed module applies: 
\begin{itemize} [leftmargin=0.5cm]
    \item MLPAE~\cite{sakurada2014anomaly} and GCNAE~\cite{kipf2017semi} are auto-encoders that serve MLP and GCN as their encoders and decoders. 
    \item DOMINANT~\cite{ding2019deep} uses a GCN encoder but reconstructs graph structure and attributes differently for anomaly detection.
    \item AnomalyDAE~\cite{fan2020anomalydae} jointly learns node attributes and topological patterns for comprehensive anomaly detection.
    \item GAAN~\cite{chen2020generative} applies generative adversarial networks to identify anomalies in graph data.
    \item CoLA~\cite{liu2022anomaly} leverages contrastive learning to capture anomaly information from multiple node pairs.
    \item GADNR~\cite{roy2024gad} focuses on neighborhood and local structure analysis to improve the detection of various anomaly types.
    \item AHFAN~\cite{wang2025graph} utilizes semantic fusion and attention mechanisms to address class and semantic inconsistency in GAD.
    \item FLAD~\cite{chen2025fiad} is a dimension-based method that injects anomalies into feature information, aiming to develop a more generalized anomaly detector.
\end{itemize}

\subsection{Comparison Results}
\par Table~\ref{tab:comparison} compares vanilla methods with their NeiGAD-enhanced versions, showing consistent ROC-AUC score improvements across all datasets and methods. 
Average performance gains for each method and dataset are provided in the last column and row, respectively. 
NeiGAD enhances all baseline methods, demonstrating its effectiveness and adaptability as a pluggable component. 
Notably, simpler models like MLPAE, GCNAE and AHFAN show larger average improvements ($5.74\%$, $4.08\%$, and $4.09\%$ respectively), while more complex methods achieve smaller yet steady gains.
NeiGAD's impact also varies across datasets. 
Improvements are more pronounced on datasets such as Books, Enron, and Pubmed (often exceeding $10\%$), whereas Cora and Citeseer show smaller gains (around $1\%$). 
This variation may stem from dataset-specific characteristics, such as graph density or feature diversity.
NeiGAD’s consistent performance across diverse architectures and datasets underscores its role as a general and effective enhancement for GAD tasks.

\begin{table}[htbp]
    \centering
    \caption{Training time for each method (s).}
    \footnotesize
    \tabcolsep=0.05cm
    \begin{tabular}{lrrrrrrrr}
        \toprule
        \textbf{Method} & \textbf{Books} & \textbf{Enron} & \textbf{Cora} & \textbf{Citeseer} & \textbf{Pubmed} & \textbf{ACM} & \textbf{Blog} & \textbf{DBLP} \\ 
        \midrule
        \multirow{2}{*}{MLPAE} & $0.41$ & $0.83$ & $0.54$ & $0.75$ & $1.26$ & $4.00$ & $1.57$ & $1.43$ \\
        & \textcolor{red}{$0$} & \textcolor{red}{$+0.01$} & \textcolor{red}{$+0.02$} & \textcolor{red}{$+0.03$} & \textcolor{red}{$+0.03$} & \textcolor{red}{$+0.38$} & \textcolor{red}{$+0.11$} & \textcolor{red}{$+0.19$} \\ 
        \midrule
        \multirow{2}{*}{GCNAE} & $0.62$ & $1.53$ & $0.98$ & $1.48$ & $2.09$ & $18.94$ & $29.26$ & $3.53$ \\
        & \textcolor{red}{$0$} & \textcolor{red}{$+0.08$} & \textcolor{red}{$+0.02$} & \textcolor{red}{$+0.02$} & \textcolor{red}{$+0.06$} & \textcolor{red}{$+0.27$} & \textcolor{red}{$+0.29$} & \textcolor{red}{$+0.13$} \\ 
        \midrule
        \multirow{2}{*}{DOM} & $0.82$ & $5.88$ & $1.30$ & $1.90$ & $10.80$ & $25.39$ & $30.24$ & $4.37$ \\
        & \textcolor{red}{$0$} & \textcolor{red}{$+0.03$} & \textcolor{red}{$+0.06$} & \textcolor{red}{$+0.20$} & \textcolor{red}{$+0.04$} & \textcolor{red}{$+0.27$} & \textcolor{red}{$+0.06$} & \textcolor{red}{$+0.05$} \\ 
        \midrule
        \multirow{2}{*}{DAE} & $0.90$ & $6.69$ & $1.13$ & $1.40$ & $11.98$ & $12.60$ & $3.91$ & $2.83$ \\
        & \textcolor{red}{$+0.02$} & \textcolor{red}{$+0.05$} & \textcolor{red}{$+0.04$} & \textcolor{red}{$+0.01$} & \textcolor{red}{$+0.03$} & \textcolor{red}{$+0.34$} & \textcolor{red}{$+0.05$} & \textcolor{red}{$+0.11$} \\ 
        \midrule
        \multirow{2}{*}{CoLA} & $0.76$ & $1.74$ & $0.96$ & $1.06$ & $1.75$ & $4.33$ & $3.01$ & $1.60$ \\
        & \textcolor{red}{$+0.05$} & \textcolor{red}{$+0.01$} & \textcolor{red}{$+0.05$} & \textcolor{red}{$+0.02$} & \textcolor{red}{$+0.03$} & \textcolor{red}{$+0.26$} & \textcolor{red}{$+0.15$} & \textcolor{red}{$+0.03$} \\ 
        \midrule
        \multirow{2}{*}{GAAN} & $1.33$ & $70.04$ & $3.33$ & $4.18$ & $50.47$ & $119.64$ & $132.40$ & $9.84$ \\
        & \textcolor{red}{$+0.10$} & \textcolor{red}{$+0.03$} & \textcolor{red}{$+0.05$} & \textcolor{red}{$+0.22$} & \textcolor{red}{$+1.24$} & \textcolor{red}{$+1.87$} & \textcolor{red}{$+1.75$} & \textcolor{red}{$+0.11$} \\ 
        \midrule
        \multirow{2}{*}{GADNR} & $6.91$ & $54.01$ & $10.21$ & $11.90$ & $59.44$ & $53.56$ & $26.01$ & $17.35$ \\
        & \textcolor{red}{$+0.03$} & \textcolor{red}{$+0.01$} & \textcolor{red}{$+0.10$} & \textcolor{red}{$+0.21$} & \textcolor{red}{$+0.14$} & \textcolor{red}{$+0.15$} & \textcolor{red}{$+0.01$} & \textcolor{red}{$+0.08$} \\
        \midrule
        \multirow{2}{*}{FIAD} & $3.55$ & $45.25$ & $6.94$ & $15.19$ & $94.96$ & $102.65$ & $23.98$ & $23.57$ \\
        & \textcolor{red}{$+0.26$} & \textcolor{red}{$+1.07$} & \textcolor{red}{$+0.06$} & \textcolor{red}{$+0.07$} & \textcolor{red}{$+1.33$} & \textcolor{red}{$+0.57$} & \textcolor{red}{$+0.81$} & \textcolor{red}{$+0.04$} \\ 
        \midrule
        \multirow{2}{*}{AHFAN} & $3.35$ & $7.12$ & $3.23$ & $3.18$ & $5.56$ & $7.50$ & $9.04$ & $3.45$ \\
        & \textcolor{red}{$+0.50$} & \textcolor{red}{$+0.12$} & \textcolor{red}{$+0.02$} & \textcolor{red}{$+0.02$} & \textcolor{red}{$+0.08$} & \textcolor{red}{$+0.02$} & \textcolor{red}{$+0.18$} & \textcolor{red}{$+0.18$} \\
        \bottomrule
    \end{tabular}
    \label{tab:training_time}
\end{table}

\subsection{Cost Comparison}
\par To evaluate scalability, we compare the training efficiency of nine baseline methods before and after integrating NeiGAD (see Table~\ref{tab:training_time}). 
The results indicate that NeiGAD is highly efficient, with the additional training time being almost negligible across all datasets. 
For most models like MLPAE and CoLA, the overhead is under $0.05$ seconds in smaller graphs. 
Notably, even for larger-scale graphs such as ACM and Blog, the relative growth in computational cost remains exceptionally low, typically within a $1$\% to $3$\% margin. 
These findings confirm that incorporating spectral neighbor information does not burden the model's training phase, making NeiGAD well-suited for real-world GAD tasks where both accuracy and speed are paramount.

\section{Conclusion}
\par NeiGAD is proposed as a novel plug-and-play module designed to enhance GAD by leveraging spectral neighbor information. 
Based on theoretical insights, eigenvectors corresponding to the largest eigenvalues of the adjacency matrix are utilized to effectively encode neighbor relationships. 
To ensure both efficiency and accuracy, $2$ to $10$ eigenvectors are selected using the ARPACK, which significantly reduces the computational complexity. 
NeiGAD offers flexibility, seamlessly integrating into various existing anomaly detection methods. Experimental results consistently demonstrate its ability to improve detection performance across diverse models and datasets. 
Future work will explore further optimizations and extend NeiGAD to handle even larger and more complex graph datasets.

\bibliographystyle{IEEEtran}
\bibliography{NeiGAD}

\end{document}